# Attention-based Memory video portrait matting

Shufeng Song[*]

[a] *Nanyang Technological University, 50 Nanyang Ave, Singapore 639798*
* e-mail: SHUFENG001@e.ntu.edu.sg

**Abstract**— We proposed a novel trimap-free video matting method based on the attention mechanism. By the nature of the problem, most existing approaches use either multiple computational expansive modules or complex algorithms to exploit temporal information fully. We designed a temporal aggregation module to compute the temporal coherence between the current frame and its two previous frames. Moreover, we also provided direct supervision for the temporal aggregation model to further boost the network's robustness. We validated our method on various testing datasets and reached state-of-the-art alpha and foreground prediction results with a low network computation complexity(9.69 GMACs for 512*512 resolution images).

## *1. INTRODUCTION*

Matting focus on using input images to generate prediction alpha and foreground with rich details. Due to its variety of applications, such as using virtual backgrounds to protect personal privacy in online meetings and editing video backgrounds for film production, matting has become a hot topic in the computer vision area.

Assuming α, F and B denote image alpha, foreground, and background, respectively. Then we can tackle the matting problem by replacing the original background B with target background B' in one linear formula:

$$I = \alpha F + (1 - \alpha)B' \qquad (1)$$

Most matting algorithms usually expected a pre-annotated trimap for each frame in the training dataset to better separate an image into alpha, foreground and unknown region[18, 30, 22]. However, manually labelling the mixed pixels occupied by alpha and foreground, which belong to the trimaps' unknown region, is impractical, especially when holes and transparent areas exist in the boundary of the matting objects[32]. Consequently, the quality of the manually annotated trimaps and the amount of dataset available are two constraints of those algorithms.

To minimise the limited precision and scale of existing trimap matting datasets, some latest work attempts to optimise the input trimap quality via trimap adaption module[6] or leverage the STM-based trimaps method to generate integral trimaps by putting several keyframes' trimaps as input[34]. Moreover, Lin *et al.* proposed Real-time high-resolution background matting[16] which puts pre-captured backgrounds as auxiliary matting input and yields excellent quality when the backgrounds images are adequately offered. Similar idea are used by Soumyadip *et al*[26]. However, it fails to deal with commonly seen dynamic backgrounds problems and using extra RGB backgrounds as input almost doubled the network parameters size.

Video matting is closely related to image matting in that each frame of the matting output essentially solves the corresponding video matting problem. Some video matting algorithms have reached a relatively good result on alpha and foreground prediction by processing video sequential frame-by-frame [16, 38, 26], despite ignoring the temporal coherence among consecutive frames. Then experiments show that fully exploiting the temporal information may improve the matting performance for several reasons. First, because the model can view successive frames and then make predictions, it can predict more coherent results and minimises flicker significantly. Second, temporal information can increase matting resilience in circumstances where an individual frame is uncertain. Third, temporal data allows the network to understand the background better over time [25]. Therefore, constructing new temporal-guided modules has become a hot topic to boost prediction accuracy.

## *2. Related Work*
### 2.1. Image segmentation

Image segmentation aims to divide an image into regions corresponding to different objects on pixels level, which could be use to separate the portrait boundary mask from the image background. State-of-the-art segmentation structure such as Atrous-Spatial-Pyramid-Pooling(ASPP) from DeepLabV3[2] has been used to help locate human subjects in matting program[16, 25]. Ke *et al.* have also modified the ASPP to lower computational overhead without decreasing the performance[19]. Our method follows the track of previous work and combines ASPP with the attention-based memory module to complete the encoding operation.

## 2.2. Attention mechanism

In short, the attention mechanism maps a query and key-value pairs to an output sum of weighted values. While weighted values are determined by their scaled-dot-product between corresponding query and keys[1]. This approach has been widely used in the segmentation area to adjust network weights dynamically[36, 37, 39]. Moreover, experiments also demonstrate fusing feature maps from successive frames via attention mechanism may improve temporal consistency[23, 31].

## 2.3. Memory network

Determining cues from which frames are worth memorizing is the central problem of the memory network. Researchers commonly use the first[15, 10] or the previous one frame[27, 11, 5] as reference. As shown in Fig. 1(a), the former structure propagates the information consecutively can handle changes well but also accelerate the spread of errors. On the other hand, the latter design in Fig. 1(b) is inefficient to extract the necessary temporal information. To take advantage of the former two approaches, we have to extend the idea of the memory network and compute the temporal coherence between the current frame and its two previous frames. Fig. 1(c) demonstrates our final memory network structure.

## 2.4. Temporal guided in video matting

Robust video matting[25] adopts the ConvGRU[20] module in the decoder at multiple scales to fully exploit the temporal information. Still, they could merely pass half of the channel to cut its computation cost through the module. Others present a complex algorithm to restore temporal backgrounds information via historical frames[14, 31]. To eliminate those issues, we propose a novel attention-based memory module with less computation complexity and reach the SOTA results.

## 3. Model Architecture

Our approach is divided into four components that jointly address the video matting task and are trained end-to-end. MobileNetV3[9] is responsible for extracting feature maps from all frames as the backbone structure. Key-value pairs are then generated in the attention-based memory block by further encoding the feature maps to measure relative matching scores over spatial-temporal locations between memory and query frames. More specifically, keys are used to determine where to reconstruct relevant memory, and values store detailed information for the alpha, foreground prediction. Finally, the temporal-guided memory module reconstructs the new feature maps. It outputs a three-channel foreground and a one-channel alpha after multiple layers up-sampling blocks and one layer output block. The overall structure is shown in Fig 2. and implementation details of each block are illustrated as follows.

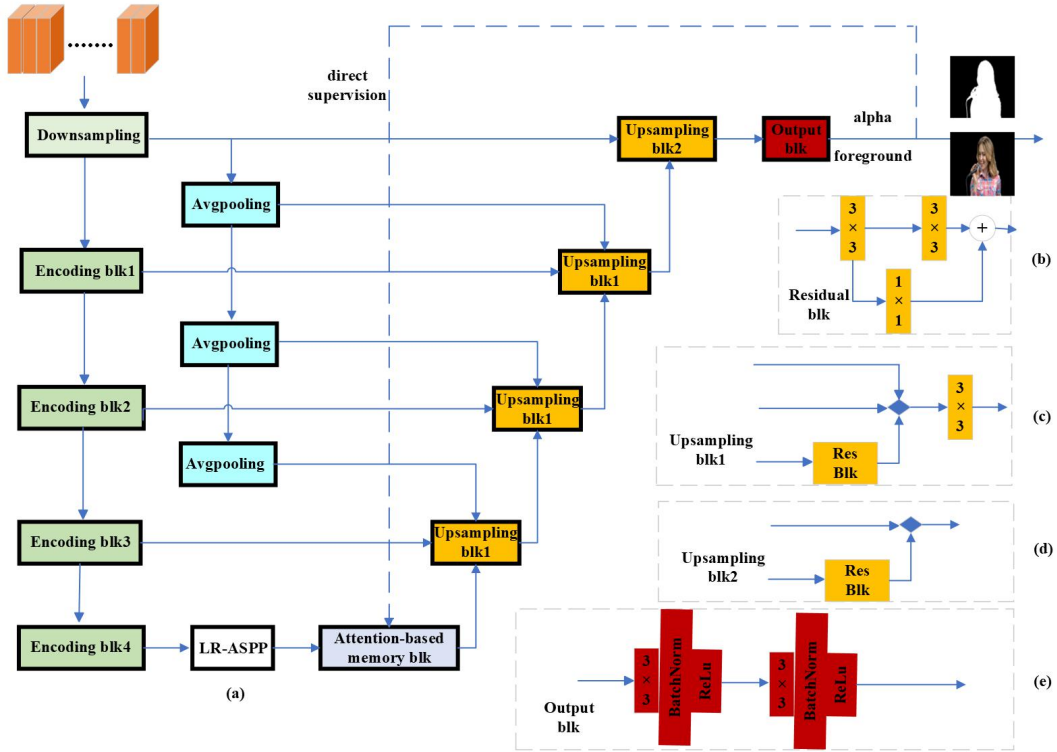

Figure 2: Overall structure of our matting network.

### 3.1. Backbone structure

MobileNetV3 pre-trained on ImageNet[24] is adopted as the default feature extraction network. Similar to most previous works, the 4th block of MobileNetV3 uses dilated convolution to preserve the feature map resolution from further down-sampling. Other complex structures such as ResNet50[8] can also be used to enhance the feature extraction quality. And eventually, feature maps at the resolution of 1/2, 1/4, 1/8, and 1/16 the down-sampled images are used for further processing.

### 3.2. Up-sampling block

Up-sampling block operates at the scale of 1/16, 1/8, 1/4, and 1/2. Fig 2. demonstrates the detailed implementation of those Up-sampling blocks. Up-sampling block 1 in Fig 2.(b) repeats three times. First, it concatenates the bilinearly up-sampled output from the previous module with the same scale feature map and down-sampled images after multiple average-pooling layers. Then a residual block is appended to reduce the channel number further and merge the feature maps. Up-sampled block 2 in Fig 2.(c) has a similar structure and function with Up-sampled block 1 but concatenates only previous output and down-sampled images before the output block.

### 3.3. Output block

Output block in Fig 2.(d) contains merely two 3×3 convolutional layers followed by Batch Normalization[12] and Relu[21] activation to downsized to 3 projection foreground channels and one alpha channel.

### 4. Matting dataset

Since most matting datasets' scale and quality limitations, we collect multiple images and video matting datasets to train our model. Following formula 1, we can generate a large set of training/validation/testing images by replacing the background of the compounded alpha and foreground images. The brief introductions of the datasets we use are listed as follows, and each matting dataset provides ground-truth foreground images and its corresponding alpha matte.

## 5. Training strategy

We followed the method of Robust video matting[25] by performing the training of semantic segmentation and matting simultaneously. On the one hand, the matting alpha is similar to the human-annotated binary mask. However, pooling and convolution operations in most segmentation tasks result in the insufficient accuracy of portrait boundary[13]. On the other hand, unlike most traditional matting algorithms in the introduction that provide trimaps or ground-truth backgrounds as an extra input, our method relies on segmentation to locate the human position and comprehend the scene semantically.

### 5.1. Training Procedure

**Image matting for stage 1:** We train our model at the resolution of 512*512 on Aisegment and Adobe image matting combination dataset for five epochs. For image matting task, exploiting time coherence among multiple independent frames is unnecessary, so T are set to 1. As for the training hyper-parameters we set the learning rate equal to 1e-4 for the backbone structure and 2e-4 for the rest network.

**Video matting for stage2:** We maximize T to our GPUs' limitation and train 20 epochs each for VideoMatte and VideoMatting108 dataset. The images resolution decrease to 512*288 while the learning rate remain unchanged from stage 1. Our network can better learn the long-term dependencies of a long video sequence to handle the real-world problem in this training stage.

**High-resolution video matting for stage3:** Deep Guided filters [7] are integrated to the end of our network in stage 3 to efficiently generate the high-resolution projection outputs by giving its final hidden feature map and corresponding low-resolution alpha and foreground. We set T=7 and trained five epochs each for VideoMatte and VideoMatting108 dataset at 1920*1080 resolution. The learning rate of Deep Guided filters is 2e−4, and the rest of the network is reduced to 1e−5 for further learning.

### 5.2. Loss function

**Image loss term $L_{im}$:** The image loss term $L_{im}$ only considers single frame prediction results, so we directly inherit loss function design from previous work. Specifically, it's a combination of alpha L1 loss $L_\alpha$, alpha Laplacian loss $L_{lap}$ [4], and composition loss $L_{com}$ [33]. Assuming $\alpha_t$, $F_t$ is the ground-truth alpha matte and foreground, then $\alpha_t'$, $F_t'$ represent their projection values. $L_{im}$ can be calculated as:

$$L_\alpha = ||\alpha_t - \alpha_t'||_1 \tag{4}$$

$$L_{lap} = \sum_s 2^{s-1}||L_{pyr}(\alpha_t) - L_{pyr}(\alpha_t')||_1 \tag{5}$$

$$L_{com} = ||\alpha_t F_t - \alpha_t F_t'||_1 \tag{6}$$

$$L_{im} = L_\alpha + L_{lap} + L_{com} \tag{7}$$

**Temporal loss term $L_t$:** Drawn inspiration from the Context Prior for Scene Segmentation[35], we use the reconstructed feature map from attention-based memory block to generate low-dimension alpha $\alpha_t^{lr}{'}$ and foreground $F_t^{lr}{'}$ for each frame and compute their L1 loss with the ground-truth alpha $\alpha_t^{lr}$ and foreground $F_t^{lr}$ that been down-sampled to the same size, which provides direct supervision to the Attention-based memory block. We also calculate the temporal coherence loss $L_{tem}$[29] to further boost the learning process.

$$L_{tem-agg} = ||\alpha_t^{lr} - \alpha_t^{lr}{'}||_1 + ||F_t^{lr} - F_t^{lr}{'}||_1 \tag{8}$$

$$L_{tem} = ||\frac{d\alpha_t}{dt} - \frac{d\alpha_t'}{dt}||_2 + ||\frac{dF_t}{dt} - \frac{dF_t'}{dt}||_2 \tag{9}$$

$$L_t = L_{tem-agg} + L_{tem} \tag{10}$$

The overall loss function $L_{vid}$ is a summation of above two different terms:

$$L_{vid} = L_{im} + L_t \tag{11}$$

## 6. Evaluation
### 6.1. Performance evaluation

Due to the property of matting tasks, we directly inherit some index values from segmentation tasks to evaluate the output results. Those index values including mean absolute difference (MAD), mean squared error (MSE), spatial gradient (Grad)[3], connectivity (Conn)[3], and dtSSD[17]. Table 1 shows the low-resolution benchmark results. The result indicates that BGM can only produce poor results with limited background information, and our method outperforms any trimap-free matting method. The lower MSE for alpha matte and foreground shows our model can generate accurate projection results. The value of dtSSD also demonstrates that the attention-based memory module in our model can fully exploit the temporal coherence. Then we run our model in 1080P, and the conclusion remains unchanged. Noted that ModNet doesn't generate foreground as its output.

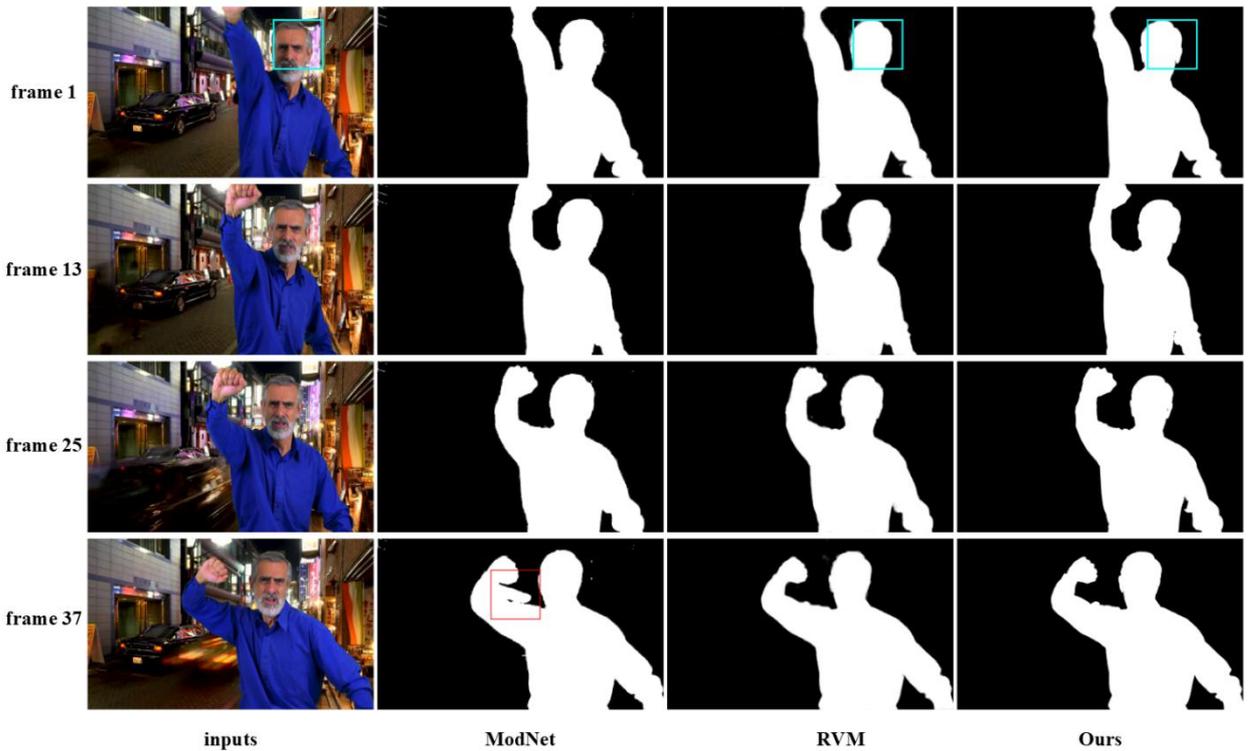

Figure 6: Display of original input and corresponding alpha projections.

## 7. CONCLUSIONS

We proposed a novel trimap-free matting structure. First, an attention-based memory module was introduced, which significantly enhanced the network's ability to exploit temporal information for videos. Secondly, we used direct supervision to train the previous block and contribute to the network robustness. Our method performs the best among the existing methods in terms of projection accuracy. We believe the proposed attention-based memory module and its training strategy have a great potential to become breakthroughs in solving the video matting problems. In the future, we plan to do more experiments on multi-object matting and look forward to the applications of matting projects.